\documentclass[10pt,twocolumn,letterpaper]{article}

\usepackage{cvpr}              

\usepackage[none]{hyphenat}

\definecolor{cvprblue}{rgb}{0.21,0.49,0.74}
\usepackage[pagebackref,breaklinks,colorlinks,allcolors=cvprblue]{hyperref}
\usepackage{url}            
\usepackage{booktabs}       
\usepackage{amsfonts}       
\usepackage{nicefrac}       
\usepackage{microtype}      
\usepackage{multirow}
\usepackage{graphicx}
\usepackage{colortbl}
\usepackage{caption}
\usepackage{amsmath}
\usepackage[accsupp]{axessibility}

\def\paperID{***} 
\def\confName{CVPR}
\def\confYear{2026}
\newcommand{\customfootnotetext}[2]{{
\renewcommand{\thefootnote}{#1}
\footnotetext[0]{#2}}}

\title{
Scene Reconstruction as Mapping Priors for 3D Detection
}

\def\authorBlock{
    Yang Fu$^{1,2}$\footnotemark[2] \quad
    Yuliang Zou$^1$ \quad
    Hao Xiang$^1$ \quad
    Xin Huang$^1$ \quad 
    Yijing Bai$^1$ \quad
    Chen Song$^1$ \quad \\
    Weijing Shi$^1$ \quad
    Govind Thattai$^1$ \quad 
    Dragomir Anguelov$^1$ \quad 
    Mingxing Tan$^1$ \quad 
    Yingwei Li$^1$ \quad \\
    \\
    $^1$Waymo LLC \qquad
    $^2$UC San Deigo \qquad 
    \\
    {\tt\small yafu@ucsd.edu} \qquad
    {\tt\small\{ylzou, ywli\}@waymo.com}
}

\author{\authorBlock}
\begin{document}
\twocolumn[{%
\renewcommand\twocolumn[1][]{#1}%
\maketitle
\begin{center}
    \centering
    \includegraphics[width=0.95\textwidth]{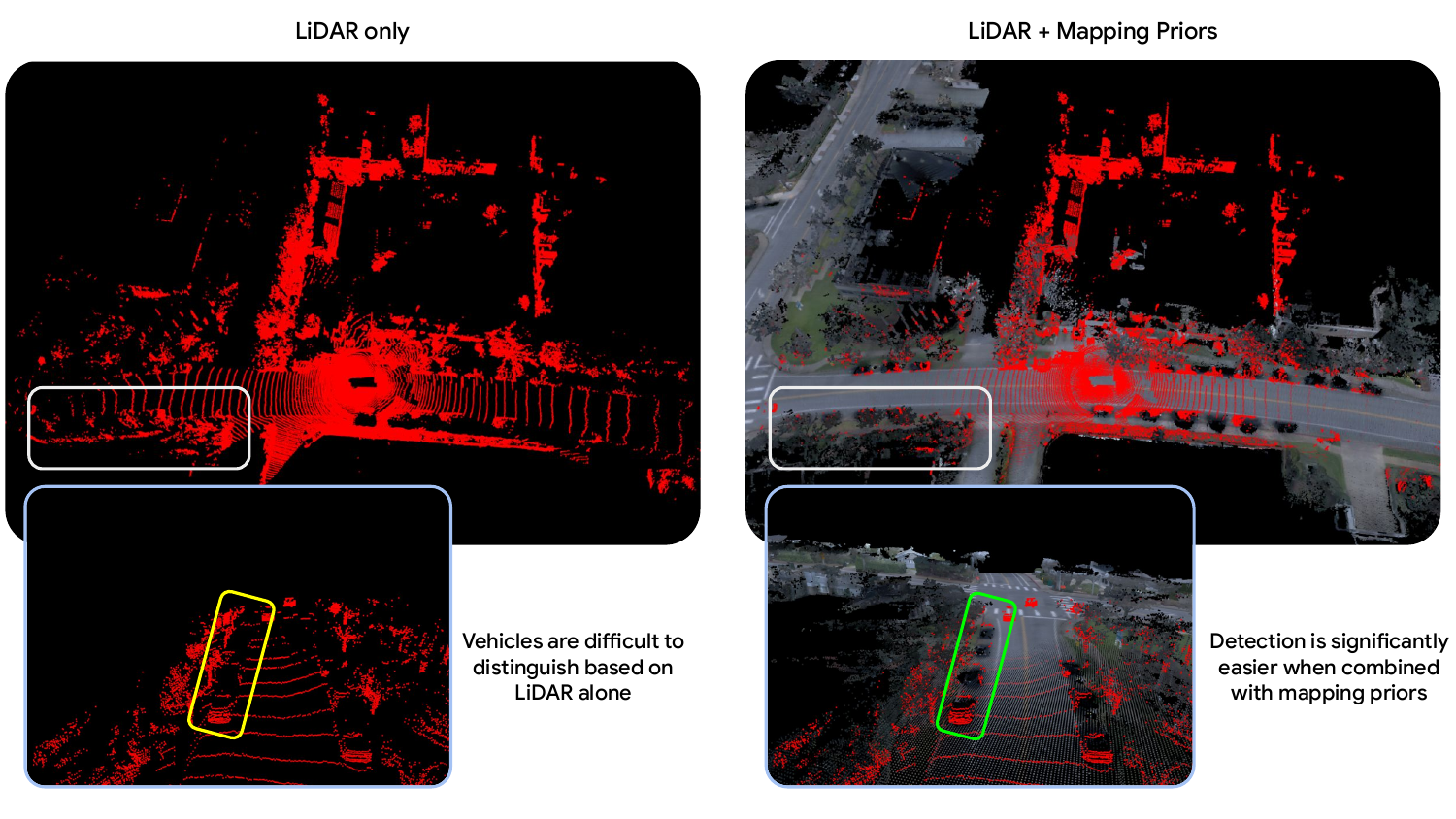}
    \vspace{-3mm}%
    \captionof{figure}{
    \textbf{Scene Reconstruction as Mapping Priors for 3D Detection.} (Left) Standard 3D detection using only sparse LiDAR data struggles to identify and distinguish multiple vehicles (yellow boxes) amidst background noise. (Right) Our method fuses the sparse LiDAR with dense, reconstructed mapping priors, providing rich context that resolves ambiguities (green boxes) and achieves accurate 3D detection.
    }
    \label{fig:teaser}
\end{center}%
}]
\customfootnotetext{$\dagger$}{This work was done during Yang Fu's internship at Waymo LLC.}
\begin{abstract}


In autonomous driving, mapping is critical for motion planning but remains an under-utilized resource for perception tasks like 3D object detection. Maps can provide robust structural priors of the static environment, suited to resolving ambiguities and correcting for sensor data sparsity or noise — issues especially prevalent for distant objects or during adverse weather conditions. However, conventional High-Definition (HD) maps are resource-intensive to obtain and maintain, which presents a challenge for achieving efficient, large-scale deployment. In this paper, we propose a scalable solution to systemically leverage mapping to improve 3D detection by overcoming two primary challenges. First, we introduce a pipeline to automatically build dense mapping priors from aggregated sensor data, eliminating the need for human labeling. Second, we design a novel \underline{\textbf{M}}apping \underline{\textbf{P}}riors \underline{\textbf{A}}ugmented \underline{\textbf{3D}} detection (MPA3D) framework to effectively integrate the mapping priors with the distinct modalities of sensor data. Our extensive experiments on the Waymo Open Dataset demonstrate that our approach achieves new state-of-the-art results, and proving the effectiveness of using scalable, reconstructed scene priors to enhance 3D detection.

\end{abstract}    
\section{Introduction}
\label{sec:intro}

Precisely recognizing and localizing objects in 3D world, also known as 3D detection, is crucial for the safety-critical L4+ autonomous driving. Due to the highest standard quality bar, it is challenging despite multiple sensors (LiDAR, camera, and radar) has been employed. However, sensors are not robust to low-visibility and low-fidelity scenarios, such as occlusions, long-range sparsity, and sensor noise due to extreme weather and lighting conditions. As described in Fig.~\ref{fig:teaser}, vehicles that are difficult to distinguish using only LiDAR data. 
Therefore, previous studies~\cite{mim, hdnet} have shown that leveraging High-Definition (HD) maps can improve 3D detection quality.

High-Definition (HD) maps, in particular, encode detailed static and semantic information, such as road layouts, lane markings, and infrastructure locations, are commonly used by L4 autonomous driving vehicles today. Several works~\cite{neuralmapprior, vectormapnet, hdmapnet, maptr, mim, hdnet} have demonstrated how HD maps can be leverage to mitigate ambiguities and noisy caused by sparsity and occlusion of the sensor data. 
However, building HD maps is not scalable as it relies heavily on manual human annotation of every road feature, making it costly to create and challenging to scale~\cite{chen2025maps,ort2022autonomous,wijaya2024high}. 

Instead of using non-scalable HD Maps, in this paper, we propose to leverage scalable reconstruction methods~\cite{3dgs,h3dgs,nerf,blocknerf,surfelgan,mcmc} as mapping priors to enhance 3D detection.
Reconstruction methods can create dense maps with photorealistic appearance and geometric information, providing a scalable and automated way to generate mapping priors directly from collected sensor data. 
We investigate two such priors: surfels~\cite{pfister2000surfels} and 3D Gaussians (3DGS)~\cite{3dgs}, which can be automatically generated from camera and LiDAR data collected during single vehicle pass or multiple traversal data. Surfel representation is simple, computationally efficient, and preserves rich geometric and appearance information. However, its primary limitation stems from a heavy reliance on LiDAR points, which can result in noisy or incomplete maps in areas of sparse data. 
On the other hand, 3DGS provides a more robust and comprehensive representation with the extra compute cost compared to surfel reconstruction. It is initialized with LiDAR data and optimized using camera data, enabling it to potentially recover regions where LiDAR data is sparse or missing. As demonstrated in Fig~\ref{fig:teaser}, with the mapping priors as the background, vehicles that are difficult to distinguish using only LiDAR data become significantly easier to detect.

To harness the complementary advantages of these priors and the raw sensor data, we employ a unified and flexible framework to integrate surfels, 3DGS, LiDAR, and camera data to leverage the complementary advantages of both mapping priors. In particular, we adopt modality-dependent encoders and dynamic voxelization~\cite{SWFormer} for feature extraction. To fuse the mapping prior information, we propose a gated fusion module to effectively modulate the influence of mapping priors while preserving the reliable LiDAR features. Moreover, we introduce a mixed-modalities training strategy to handle arbitrary combinations of these modalities at both training and inference time. Our contributions can be summarized as follows:
\begin{itemize}
    \item We introduce a scalable data pipeline to generate mapping priors, \eg, surfel and 3DGS, across thousands of scenes.
    \item We develop a unified framework with a gated fusion mechanism that flexibly integrates surfels, 3DGS, LiDAR, and camera data. A mix-training strategy is proposed to support arbitrary combinations of these modalities, achieving robust performance across scenarios with varied access to these data sources.
    \item We conduct extensive experiments on the Waymo Open Dataset~\cite{Waymo} to demonstrate state-of-the-art performance and validate the effectiveness of each component.
\end{itemize}

\section{Related Work}
\label{sec:related_work}
\paragraph{3D Object Detection.} 
The task of 3D object detection is foundational for autonomous driving. Modern 3D object detection approaches can be broadly categorized by input modalities and representations. LiDAR-based detectors~\cite{ku2018joint,simony2018complex,PointPillar,li20173d,SWFormer,SECOND,HEDNet,CenterFormer,PointTransformer, Safdnet} have achieved dominant performance by voxelizing point clouds and applying sparse convolutional backbones. For instance, SAFDNet~\cite{Safdnet} forgoes dense Bird’s-Eye-View (BEV) maps and instead introduces a fully sparse one-stage detector with an adaptive feature diffusion mechanism. By eliminating the dense mapping bottleneck, SAFDNet achieves competitive accuracy with significantly reduced latency.
Concurrently, camera-only detection~\cite{bevformer,FCOS3D,bevmap} underwent a critical paradigm shift. Early 2.5D methods like FCOS3D~\cite{FCOS3D} regressed 3D properties from the 2D perspective view. The field then pivoted to BEV-based representations, which provide a unified space for multi-modal fusion. This view transformation is achieved either explicitly, as in Lift-Splat-Shoot (LSS)~\cite{lss}, which predicts a categorical depth distribution, or implicitly, as in BEVFormer~\cite{bevformer}, which uses transformer-based spatial cross-attention to project 2D features into a BEV grid. 
Our approach is built upon SWFormer~\cite{SWFormer} and integrates camera features using the LSS module to lift and fuse them into a shared BEV space.

\paragraph{LiDAR Long-Context Temporal Fusion.}
Long-context temporal fusion addresses the inherent limitations of single-frame perception, such as occlusion and sparsity. A common solution is to align multi-frame data by transforming LiDAR points from previous frames into the current frame's coordinate system and then concatenating all multi-frame LiDAR data~\cite{strobe, lef, bevformer, CenterPoint, hindsight, videobev}. For instance, VideoBEV~\cite{videobev} maintains a dense, holistic BEV feature map and updates it via a simpler, decoupled recurrent module.
Although these multi-frame techniques achieve better detection performance by providing a denser representation of the surrounding environment, they cannot handle long sequences due to computational and memory constraints. 
In contrast, the object-based paradigm is more efficient by focusing only on foreground proposals. This category includes ``detect-track-fuse" methods~\cite{Mppnet, Modar, pnpnet, motiontrack}, which associate previous detections over time via a explicit tracker. However, the primary drawback of this sequential approach is that tracking failures (\eg, incorrect associations or ID switches) propagate and accumulate. This error accumulation directly degrades the quality of the aggregated temporal features, leading to reduced detection performance. 
To alleviate the sensitivity to 3D proposals and tracking, other methods~\cite{Msf, 3dman,hou2023query,MAD} employ attention modules to fuse the 3D detections from the current frame with proposals from past frames, which are represented by previous detections or explicit trajectory forecasts. For example, MAD~\cite{MAD} maintains an "object memory bank" and utilizes trajectory forecasts to align past proposals to the current time. This allows the model to recover objects from memory even if the current detector produces a false negative.

In contrast to these long-context temporal fusion approaches, our method proposes an alternative paradigm by leveraging mapping priors (\eg, surfels and 3DGS) that provide rich contextual information from pre-built scene reconstructions. This approach achieves comparable or superior performance without requiring long-term frame aggregation or explicit tracking.

\begin{figure*}[tp]
    \centering
    \includegraphics[width=1.\textwidth]{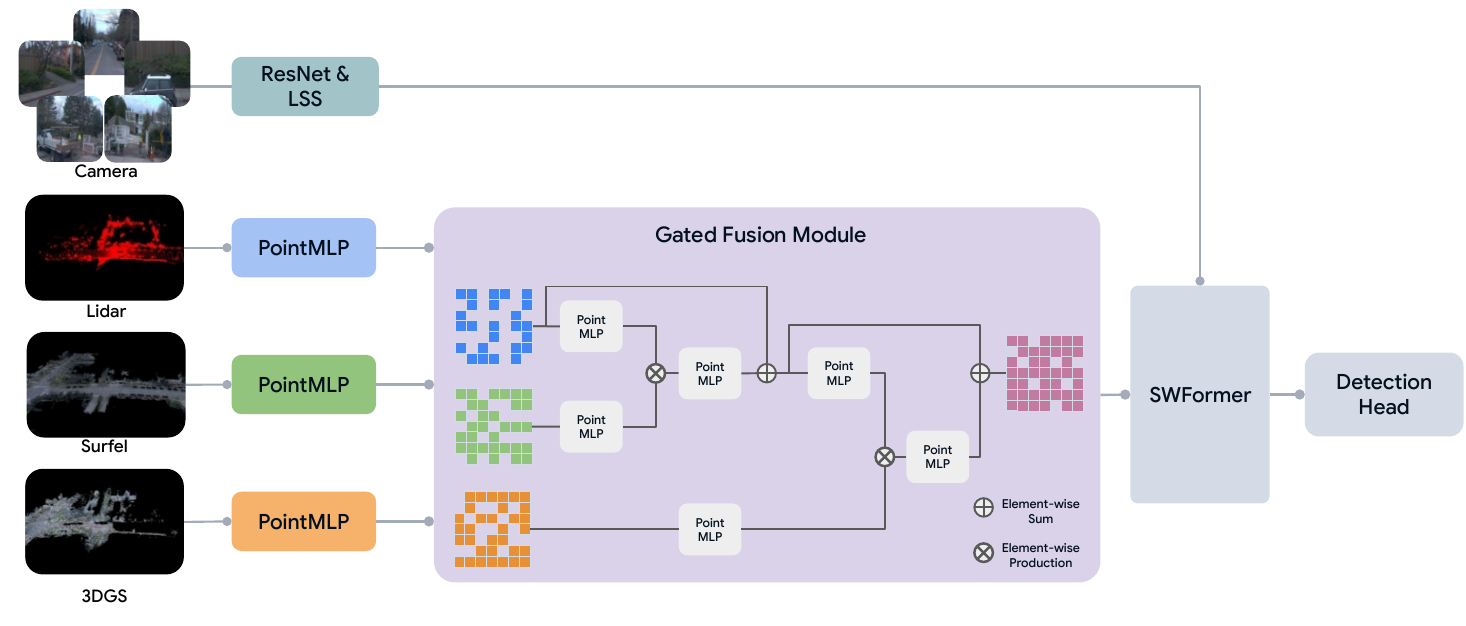}
    \vspace{-8mm}
    \caption{ 
    \textbf{Overview of \textit{MPA3D}.} The framework processes multi-view camera images via a ResNet backbone and Lift-Splat-Shoot~\cite{lss} to generate BEV representations. Simultaneously, LiDAR point clouds and mapping priors (\eg, surfels and 3DGS) are encoded using PointMLPs. A novel Gated Fusion Module adaptively integrates these sparse 3D modalities through hierarchical, element-wise gating operations. Finally, the fused geometric features are combined with the image representations and processed by an SWFormer backbone and a detection head to yield 3D bounding box predictions.}
    \label{fig:method}
    \vspace{-4mm}
\end{figure*}

\paragraph{Mapping for Autonomous Driving.}
High-Definition (HD) maps serve as crucial priors for autonomous driving, providing centimeter-level road geometry and semantic structure that enhance perception, localization, and motion planning. However, conventional HD maps are typically generated through offline LiDAR surveys and extensive manual labeling, making them costly, static, and difficult to maintain across large regions. To alleviate these limitations, recent works have explored learning-based and online mapping approaches. HDMapNet~\cite{hdmapnet} introduced an end-to-end framework for generating dense semantic maps from multi-modal sensor data, establishing a foundation for automatic HD map construction. Vectorized map prediction methods such as VectorMapNet~\cite{vectormapnet} and MapTR~\cite{maptr} further advanced this paradigm by representing road elements as structured polylines decoded by neural networks, enabling real-time, fine-grained map generation. NeuralMapPrior~\cite{neuralmapprior} extended this idea by maintaining a global latent representation that can be incrementally updated from fleet observations. Despite these advances, most existing methods still depend on high-quality HD maps or densely annotated training data, which constrains their scalability. Therefore, we propose a scalable mapping prior that can be automatically built from aggregated sensor data without human annotation, providing a lightweight yet robust structural prior to improve 3D perception under diverse driving environments.

\section{Method}\label{sec:method}

In this section, we describe MPA3D, \underline{\textbf{M}}apping \underline{\textbf{P}}riors \underline{\textbf{A}}ugmented \underline{\textbf{3D}} object detection. The overall framework is shown in Figure~\ref{fig:method}. Built upon~\cite{SWFormer}, we take two types of mapping priors, \eg, surfels and 3DGS, as additional inputs and employ a gated fusion strategy to integrate them with LiDAR and camera data. In the following subsections, we first introduce a scalable data pipeline to generate huge amount of surfels and 3DGS without any human annotations. We then demonstrate how to integrate multi-sensor data with mapping priors.

\subsection{Scalable Mapping Prior Generation}\label{sec3.1}
\paragraph{Surfel Map Reconstruction.} 
Following~\cite{surfelgan}, we generate surfel map with faithful preservation of sensor information while maintaining low computational and storage costs.
Specifically, we discretize the scene into a 3D voxel grid with fixed voxel size of 0.25m and process multi-traversal LiDAR scan data. For each voxel, we construct a surfel disk~\cite{pfister2000surfels} by estimating the mean coordinate, surface normal, and mean color from its binned LiDAR points and corresponding camera pixels. As a result, we represent a scene with a set of surfel disks: $\mathcal{S}=\{x_i, n_i, c_i\}_{i=1:N_\text{S}}$, where $x$, $n$ and $c$ refer to the surfel position, normal and color, respectively and $N_\text{S}$ is the numberf of surfel disk.
Since the creation of each surfel is relatively independent, we can partition a large region into multiple sub-regions and conduct reconstruction with massive parallelism, which allows us easily scale up surfel map creation to multiple cities.
%

\paragraph{3DGS Map Reconstruction.} 
Surfel maps are heavily reliant on LiDAR data, making them susceptible to noise and artifacts from sensor perturbations. To address this, we propose to use 3DGS as a complementary mapping prior. Specifically, a 3DGS scene is represented by a set of 3D Gaussians:$$\boldsymbol{\mathcal{G}} = \{(\boldsymbol{\mu}, \boldsymbol{\text{SH}}, \boldsymbol{r}, \boldsymbol{s}, \alpha)_i\}_{i=1:N_\text{G}}$$where each Gaussian $i$ is defined by its 3D position $\boldsymbol{\mu}$, spherical harmonics (SH) coefficients $\boldsymbol{\text{SH}}$, quaternion rotation $\boldsymbol{r}$, scale $\boldsymbol{s}$, opacity $\alpha$, and $N_\text{G}$ is the total number of Gaussians. We initialize the Gaussian positions $\boldsymbol{\mu}$ from LiDAR data, but subsequently optimize all attributes (including position) by minimizing a photometric loss against the camera images. This is a key difference from surfel maps, whose disk positions are fixed by the initial LiDAR points. By refining the Gaussian positions during optimization, 3DGS can effectively correct for noise and densify regions with sparse or incomplete LiDAR data.
Note that we implemented custom CUDA kernels to perform 3D Gaussian ray-tracing (similar to \cite{wu20253dgut}) to better represent fine geometry while still maintain compute efficiency, instead of the original splatting implementation~\cite{3dgs}.

\paragraph{Dynamic Object Removal.} Since 3DGS reconstruction assumes a static scene, we must first remove dynamic objects.
For the training set, we utilize the 3D bounding box annotations to remove dynamic points during initialization and project object-specific LiDAR points into 2D masks. These masks are then used to exclude moving objects during 3DGS optimization.
For inference, we first run our model in its map-free configuration (enabled by our mix-training strategy, Sec.~\ref{sec3.3}) to predict initial 3D bounding boxes. These predictions are then used to create the necessary masks for on-the-fly map generation.
We leave the accurate dynamic reconstruction to future work.


\subsection{Mapping Prior Augmented 3D Detection}\label{sec3.2}
Our model architecture, illustrated in Fig.~\ref{fig:method}, processes camera data $I$, LiDAR data $P$, and mapping priors $M$. First, modality-dependent encoders extract features, which are then projected into a shared Bird’s-Eye-View (BEV) space, similar to~\cite{bevfusion}. These BEV features are subsequently processed by our proposed gated fusion module, which leverages the mapping priors to enhance the sensor-derived features. Finally, the fused BEV representation is fed into a sparse-window transformer~\cite{SWFormer} and a detection head to produce the final 3D object detections.

\paragraph{Modality-Dependent Feature Extraction.} We employ different encoders tailored to each input modality after transforming it to the same vehicle coordinate. For LiDAR points, we follow~\cite{SWFormer} to extract point-level features into the BEV space: raw points are first grouped into voxels via dynamic voxelization~\cite{dynvoxel}, then processed through PointMLP~\cite{pointmlp} to generate point-wise features, which are subsequently aggregated into BEV features. For camera data, we use the lift-splat-shoot (LSS)~\cite{lss} module to lift image features into 3D frustums and splat them into the shared BEV space. More architecture details about LiDAR and camera feature extraction can be found in~\cite{li2026stellar}. For surfel points, we treat surfel disk centers as pseudo-LiDAR points and apply the similar encoder architecture as LiDAR encoder, leveraging their geometric properties. For 3DGS, we convert each Gaussian into a point-like representation characterized by all Gaussian attributes. Notably, we only utilize the first order of SH coefficient and adopt the 6D continuous rotation representation~\cite{6drotation} instead of quaternions to encode colors and orientations. We introduce an additional projection layer for each modalities to align them into the same embedding space. As a result, we have a set of LiDAR, camera, surfel and 3DGS features denoted as $\{f_{\text{lidar}}\in\mathbb{R}^{N_\text{lidar}\times d}, f_{\text{camera}}\in\mathbb{R}^{N_\text{camera}\times d}, f_{\text{surfel}}\in\mathbb{R}^{N_\text{surfel}\times d}, f_{\text{Gaussian}}\in\mathbb{R}^{N_\text{Gaussian}\times d}\}$, where $N_{*}$ refers to the number of voxels after dynamic voxelization.

\paragraph{Feature Enhancement via Gated Fusion.} Since we employ the same dynamic voxelization under the same vehicle coordinate for LiDAR, surfel and 3DGS inputs, the resulting lidar feature $f_{\text{lidar}}$, surfel feature $f_{\text{surfel}}$ and Gaussian feature $f_{\text{gaussian}}$ share the same BEV grid structure with $N_\text{lidar}=N_\text{surfel}=N_\text{gaussian}$. A straightforward baseline is to average features from different modalities that occupy the same voxel using a segment mean operation over the sparse feature tensor. However, this naive averaging approach has a critical limitation: it is biased toward the modality with the highest point density. For instance, if a voxel contains 95 LiDAR points and only 5 Gaussians, the averaged feature is dominated by LiDAR (95\%), effectively marginalizing the complementary information from Gaussians. This density imbalance prevents effective multi-modal fusion and fails to leverage the unique strengths of each representation. 

To mitigate this issue, we propose a gated fusion module to learn adaptive weighting based on local scene characteristics. As described in Fig.~\ref{fig:method}, we first aggregate features within a single voxel from different modalities independently using segment mean operation, yielding modality-level features $\bar{f}_{\text{lidar}}$, $\bar{f}_{\text{surfel}}$, and $\bar{f}_{\text{gaussian}}$. 
We treat the LiDAR feature $\bar{f}_{\text{lidar}}$ as the primary feature, which acts as a gate for the mapping priors. 
We employ several PointMLPs for internal feature transformations and for output projections, denoted as $\sigma_{*} \in \mathbb{R^{d\times d}}$ and $\phi_{*}\in\mathbb{R}^{d \times d}$.
First, we compute the gated contribution from the surfel map:
$$\alpha_{\text{surfel}} = \boldsymbol{\text{Swish}}(\sigma_{\text{in}}(\bar{f}_{\text{lidar}})) \cdot \sigma_{\text{surfel}}(\bar{f}_{\text{surfel}})$$
Here, $\alpha_{\text{surfel}}$ represents the modulated feature contribution of the surfel map, informed by the LiDAR context and $\text{Swish}(\cdot)$ stands for Swish activation function~\cite{hendrycks2016gaussian}.
The intermediate fused feature is then computed via a residual connection:
$$f_{\text{inter}} = \phi_{\text{surfel}}(\alpha_\text{surfel}) + \bar{f}_\text{lidar}$$
We then repeat this gated process for the 3D Gaussian prior, using the intermediate fused feature as the new gate:
$$\alpha_{\text{Gaussian}} = \boldsymbol{\text{Swish}}(\sigma_{\text{inter}}(f_{\text{inter}})) \cdot \sigma_{\text{Gaussian}}(\bar{f}_{\text{Gaussian}})$$
The final fused feature $f_{\text{fused}}$ is obtained by combining this new contribution with the intermediate feature:
$$f_{\text{fused}} = \phi_{\text{Gaussian}}(\alpha_{\text{Gaussian}}) + f_{\text{inter}}$$
This design allows the network to adaptively modulate the influence of mapping priors while preserving the reliable LiDAR features through the skip connection. Finally, $f_{\text{fused}}$ is concatenated with the dense camera feature $f_{\text{camera}}$ through $f_{\text{final}}=[f_{\text{camera}}, f_{\text{fused}}] \in \mathbb{R}^{(N_\text{camera}+N_\text{lidar})\times d}$. This combined BEV representation serves as the input to the subsequent sparse-window transformer.

\subsection{Mixed-Modality Training Strategy}\label{sec3.3}
As previously discussed, our pipeline can effectively integrate and fuse mapping priors with common sensor data. However, in practice, the availability of these mapping priors cannot be guaranteed: Surfel maps and 3D Gaussian maps may be unavailable due to reconstruction failures caused by insufficient data coverage or challenging environmental conditions. Moreover, during inference, we assume no prior knowledge about 3D bounding boxes is available. Therefore, our model must robustly handle arbitrary combinations of input modalities without relying on complete mapping priors or pre-existing annotations.

To address this challenge, we propose a mix-training strategy that enables the model to operate effectively under varying modality availability. During training, we randomly sample different combinations of input modalities for each training sample, simulating diverse real-world scenarios. In particular, we assume the sensor data, \eg, camera and LiDAR, always exist and define the following modality combinations: sensor + surfel, sensor + 3DGS, sensor + surfel + 3DGS. 
For each training batch, we randomly drop surfel or 3DGS with predefined probabilities. When a modality is dropped, its corresponding feature is set to zero, and the gated fusion module learns to automatically suppress its contribution through the attention weights. The gated fusion module enables this flexibility through three key design choices. First, the modality-independent aggregation allows missing modalities to contribute zero features without architectural changes. Second, the learnable attention weights automatically suppress contributions from absent or unreliable modalities; for example, the network learns to set $\alpha_{\text{surfel}} \approx 0$ when $\bar{f}_{\text{surfel}} = \mathbf{0}$. Third, the residual connection with the LiDAR feature $\bar{f}_{\text{lidar}}$ ensures stable performance. If all mapping priors are absent, the output $f_{\text{fused}}$ naturally defaults to $\bar{f}_{\text{lidar}}$. This design handles any modality combination without requiring explicit masking or conditional branches.
This mix-training strategy forces the network to learn robust feature representations that do not over-rely on any single modality.
During inference, the model seamlessly adapts to whatever modalities are available without requiring architectural modifications or retraining. 

\begin{table*}[ht]
\centering
\resizebox{\textwidth}{!}{%
\begin{tabular}{lcccccccccccccccc}
\toprule
& \multicolumn{2}{c}{Overall L1} & \multicolumn{2}{c}{Overall L2} & \multicolumn{2}{c}{Vehicle L1} & \multicolumn{2}{c}{Vehicle L2} & \multicolumn{2}{c}{Pedestrian L1} & \multicolumn{2}{c}{Pedestrian L2} & \multicolumn{2}{c}{Cyclist L1} & \multicolumn{2}{c}{Cyclist L2} \\
\cmidrule(lr){2-3} \cmidrule(lr){4-5} \cmidrule(lr){6-7} \cmidrule(lr){8-9} \cmidrule(lr){10-11} \cmidrule(lr){12-13} \cmidrule(lr){14-15} \cmidrule(lr){16-17}
Method & AP & APH & AP & APH & AP & APH & AP & APH & AP & APH & AP & APH & AP & APH & AP & APH \\
\midrule
SECOND~\cite{SECOND} & 67.2 & 63.1 & 61.0 & 57.2 & 72.3 & 71.7 & 63.9 & 63.3 & 68.7 & 58.2 & 60.7 & 51.3 & 60.6 & 59.3 & 58.3 & 57.0\\
PointPillar~\cite{PointPillar} & 69.0 & 63.5 & 62.8 & 57.8 & 72.1 & 71.5 & 63.6 & 63.1 & 70.6 & 56.7 & 62.8 & 50.3 & 64.4 & 62.3 & 61.9 & 59.9 \\
Part-A2-Net~\cite{PartA2Net} & 73.6 & 70.3 & 66.9 & 63.8 & 77.1 & 76.5 & 68.5 & 68.0 & 75.2 & 66.9 & 66.2 & 58.6 & 68.6 & 67.4 & 66.1 & 64.9 \\
SST~\cite{SST} & 74.5 & 71.0 & 67.8 & 64.6 & 74.2 & 73.8 & 65.5 & 65.1 & 78.7 & 69.6 & 70.0 & 61.7 & 70.7 & 69.6 & 68.0 & 66.9 \\
Centerpoint~\cite{CenterPoint} & 76.1 & 73.5 & 70.0 & 67.6 & 75.7 & 75.2 & 67.9 & 67.4 & 77.6 & 71.6 & 70.1 & 64.4 & 74.9 & 73.8 & 72.1 & 71.0 \\
CenterFormer~\cite{CenterFormer} & 75.6 & 73.2 & 71.4 & 69.1 & 75.0 & 74.4 & 69.9 & 69.4 & 78.0 & 72.4 & 73.1 & 67.7 & 73.8 & 72.7 & 71.3 & 70.2\\
PillarNet-34~\cite{PillarNet} & 77.3 & 74.6 & 71.0 & 68.5 & 79.1 & 78.6 & 70.9 & 70.5 & 80.6 & 74.0 & 72.3 & 66.2 & 72.3 & 71.2 & 69.7 & 68.7 \\
PV-RCNN++~\cite{PV-RCNN++} & 78.1 & 75.9 & 71.7 & 69.5 & 79.3 & 78.8 & 70.6 & 70.2 & 81.3 & 76.3 & 73.2 & 68.0 & 73.7 & 72.7 & 71.2 & 70.2 \\
DSVT-Voxel~\cite{DSVT} & 80.3 & 78.2 & 74.0 & 72.1 & 79.7 & 79.3 & 71.4 & 71.0 & 83.7 & 78.9 & 76.1 & 71.5 & 77.5 & 76.5 & 74.6 & 73.7 \\
SWFormer~\cite{SWFormer} & - & - & - & - & 77.8 & 77.3  & 69.2 & 68.8 & 80.9 & 72.7 & 72.5 & 64.9 & - & - & - & - \\
FSDv1~\cite{FSD} & 79.6 & 77.4 & 72.9 & 70.8 & 79.2 & 78.8 & 70.5 & 70.1 & 82.6 & 77.3 & 73.9 & 69.1 & 77.1 & 76.0 & 74.4 & 73.3 \\
FSDv2~\cite{FSDv2} & 81.8 & 79.5 & 75.6 & 73.5 & 79.8 & 79.3 & 71.4 & 71.0 & 84.8 & 79.7 & 77.4 & 72.5 & 80.7 & 79.6 & 77.9 & 76.8 \\
HEDNet~\cite{HEDNet} & 81.6 & 79.7 & 75.6 & 73.7 & 80.9 & 80.5 & 73.1 & 72.7 & 84.6 & 80.2 & 77.1 & 72.8 & 79.4 & 78.5 & 76.6 & 75.6 \\
SAFDNet~\cite{Safdnet} & 81.7 & 79.7 & 75.5 & 73.6 & 80.5 & 80.0 & 72.5 & 72.1 & 84.7 & 80.2 & 77.1 & 72.9 & 79.8 & 78.8 & 76.9 & 75.9 \\

\midrule
Centerpoint 2f~\cite{CenterPoint} & 77.5 & 75.8 & 71.7 & 70.1 & 76.4 & 75.9 & 68.7 & 68.2 & 79.2 & 75.6 & 71.9 & 68.5 & 76.8 & 75.9 & 74.4 & 73.5 \\
HEDNet 4f~\cite{HEDNet} & 83.6 & 82.3 & 78.1 & 76.8 & 82.4 & 81.9 & 75.1 & 74.6 & 86.3 & 83.6 & 79.4 & 76.8 & 82.2 & 81.4 & 79.9 & 79.1 \\
SAFDNet 4f~\cite{Safdnet} & 83.9 & 82.6 & 78.4 & 77.1 & 82.8 & 82.3 & 75.4 & 74.9 & 86.8 & 84.2 & 80.1 & 77.5 & 82.0 & 81.1 & 79.6 & 78.8 \\
\rowcolor{blue!10}
MPA3D (Ours) 4f & \textbf{86.4} & \textbf{84.9} & \textbf{81.6} & \textbf{80.1} & \textbf{85.4} & \textbf{84.9} & \textbf{78.7} & \textbf{78.2} & \textbf{87.1} & \textbf{84.1} & \textbf{81.2} & \textbf{78.2} & \textbf{86.7} & \textbf{85.8} & \textbf{84.9} & \textbf{84.0} \\
\bottomrule
\end{tabular}
}
\caption{\textbf{Performance comparisons on WOD validation set.} Note that all methods listed utilize single frame unless specified by ``xf". 
We use \textbf{bold numbers} to highlight the best results.}
\vspace{-3mm}
\label{tab:wod_val}
\end{table*}
\subsection{Training Objectives}
We follow SWFormer~\cite{SWFormer} to regress 3D bounding boxes using a combination of heatmap, bounding box, and foreground segmentation losses. Specifically, the heatmap loss $L_{\text{hm}}^c$ is computed as a penalty-reduced focal loss~\cite{Centernet, lin2017focal} for each object class $c$. For 3D bounding box regression, we parameterize boxes as $\boldsymbol{b}=\{d_x, d_y, d_z, l, w, h, \theta\}$, where $\{d_x, d_y, d_z\}$ are center offsets relative to voxel centers, and $\{l, w, h, \theta\}$ represent dimensions and heading. Following~\cite{sun2021rsn}, the bounding box loss $L_{\text{bbox}}^c$ comprises a bin loss for heading, a Smooth L1 loss for other parameters, and an IoU loss~\cite{zhou2019iou} to improve overall accuracy. 
Additionally, we jointly train a class-aware foreground segmentation task, where each voxel is supervised by a binary focal loss $L_{\text{seg}}^c$ for each class. The total loss is defined as:
$$
L = \sum_c (\lambda_{\text{hm}} L_{\text{hm}}^c + \lambda_{\text{bbox}} L_{\text{bbox}}^c + \lambda_{\text{seg}} L_{\text{seg}}^c)
$$
where the loss weights are set to $\lambda_{\text{hm}} = 1.0$, $\lambda_{\text{bbox}} = 2.0$, and $\lambda_{\text{seg}} = 1.0$ during training.

\section{Experiments}\label{sec:experiments}
\subsection{Datasets and Metrics}
We conduct experiments and ablation studies mainly on the Waymo Open Dataset (WOD)~\cite{Waymo}. The dataset contains 1,150 scenes, split into 798 training scenes, 202 validation scenes, and 150 test scenes. Each scene contains approximately 200 frames, where each frame captures a full 360-degree view around the ego-vehicle.
We produce surfel map and 3DGS map for each scene based on LiDAR and camera data from WOD or in-house sensor data.
Additionally, we collect around 7M additional sequences for pretraining, and 600k of them have map prior data. These additional sequences enable us to pre-train the mapping prior encoders on large-scale diverse scenarios, improving their feature extraction capabilities before fine-tuning on WOD.

The evaluation metrics include mean average precision (mAP) and mAP weighted by heading accuracy (mAPH), computed at 3D intersection-over-union (IoU) thresholds of 0.5, 0.5 and 0.7 for pedestrian, cyclist and vechlie, respectively. The metrics are further broken down into two difficulty levels: LEVEL\_1 (L1) for objects with more than five LiDAR points and are not marked as ``hard", and LEVEL\_2 (L2) for objects with at least one LiDAR point. Following common practice~\cite{SWFormer}, the detection range is limited to 75 meters. 

\subsection{Implementation Details}
\paragraph{Model Architecture and Training.} During training, we ignore all ground truth boxes with fewer than five points inside. We set the voxel size of dynamic voxelization to $0.2$m and the maximum voxel number to $250$K. The SWFormer architecture consists of five transformer blocks with $[4, 6, 4, 6, 4]$ layers respectively. Each transformer layer has $256$ channels, $8$ attention heads, and an MLP expansion ratio of $2$. The model is trained end-to-end
using $256$ TPU cores with the LAMB optimizer~\cite{Lamb} for $20$K steps. We use a cosine learning rate schedule starting from $1\text{e}^{-5}$ and decaying to zero until the end of training. 
\paragraph{Data augmentation.} We adopt several 3D data augmentation techniques described in~\cite{lidaraugment}. In particular, we apply random rotation with yaw angle uniformly sampled from $[-\pi, \pi]$ with probability of $0.74$, random flipping along y-axis with probability of $0.5$, random scaling with scaling factor uniformly chosen within $[0.95, 1.05)$ and random point dropout with probability of $0.05$. 
To ensure the same augmentations are consistently applied to the camera data, instead of transforming the camera extrinsics and intrinsics, we directly transform the points with camera features lifted from Lift-Splat-Shoot (LSS)~\cite{lss}. For mapping priors (surfel and 3DGS), we apply the same spatial transformations to their coordinates and orientations to preserve alignment across all modalities.

\paragraph{Efficiency of scalable mapping prior generation.}
To optimize mapping prior generation throughput, we implement scalable pipelines for both surfel and 3DGS using the Apache Beam API in a MapReduce framework.
Using this framework, we can generate both surfel and 3DGS maps for 600,000 scenes within 10 days with massive parallesim, using thousands of CPU cores.
This automated generation process represents an affordable computational budget compared to manual annotation of HD maps, which typically requires extensive human labor.

\begin{table}[t]
\centering
\resizebox{\columnwidth}{!}{
\begin{tabular}{clcccc}
\toprule
& Method & AP L1 & APH L1 & AP L2 & APH L2 \\
\midrule
\multirow{6}{*}{\rotatebox{90}{\textit{Validation}}}
& LEF~\cite{lef}  & 79.6 & 79.2 & 71.4 & 70.9 \\
& MoDAR~\cite{Modar}  & - & - & - & 72.5 \\
& MPPNet~\cite{Mppnet}  & 81.6 & 81.1 & 76.0 & 74.8 \\
& MSF~\cite{Msf}  & 82.2 & 80.7 & 76.8 & 75.5 \\
& PTT~\cite{Ptt}  & 82.7 & 80.7 & 77.7 & 75.7 \\
& MAD~\cite{MAD}  & 85.8 & 84.2 & 81.0 & 79.4 \\
\rowcolor{blue!10} 
& MPA3D (Ours) & \textbf{86.4} & \textbf{84.9} & \textbf{81.6} & \textbf{80.1} \\
\midrule
\multirow{4}{*}{\rotatebox{90}{\textit{Testing}}}
& 3D-MAN~\cite{3dman}  & 49.6 & 48.1 & 44.8 & 43.4 \\
& MPPNet~\cite{Mppnet}  & 81.8 & 80.6 & 76.9 & 75.7 \\
& MSF~\cite{Msf}  & 83.1 & 81.7 & 78.3 & 77.0 \\
& MAD~\cite{MAD}  & 86.0 & 84.3 & 81.8 & 80.2 \\
\rowcolor{blue!10}
& MPA3D (Ours) & \textbf{87.2} & \textbf{85.9} & \textbf{83.0} & \textbf{81.6} \\
\bottomrule
\end{tabular}
}
\caption{\textbf{Comparisons with temporal fusion based approaches.}
We use \textbf{bold numbers} to highlight the best results.}
\label{tab:wod_fusion}
\end{table}
\subsection{Main Results}
\paragraph{Comparison with state-of-the-art works.}

We report the performance of our \textsc{MPA3D} on the WOD validation split against off-the-shelf 3D detectors in Table~\ref{tab:wod_val}. We categorize existing approaches into two groups. The first group contains methods that perform 3D detection using only the current frame, while the second group consists of methods that use multiple-frame inputs. These multi-frame methods extend their single-frame counterparts by concatenating sensor data from previous frames and usually achieve superior performance compared to single-frame methods. Using four-frame inputs, our approach demonstrates superior performance over all existing approaches in the multi-frame category. For instance, we outperform the previous best multi-frame method SAFDNet~\cite{Safdnet} by $2.2\%$ in overall L1 APH and $2.7\%$ in overall L2 APH.

Additionally, we compare our approach with existing temporal-fusion-based methods in Table~\ref{tab:wod_fusion}. Notably, these approaches utilize up to 99 previous frames of sensor data, which is substantially more than our four-frame input. Compared to the best temporal fusion method MAD~\cite{MAD}, our approach improves the overall L2 AP and APH by $0.2\%$ and $0.4\%$ on validation set and $0.9\%$ and $1.2\%$ on testing set.

Furthermore, Table~\ref{tab:wod_test} reports our state-of-the-art performance on the WOD test set leaderboard among all online methods that do not use ensembles or test-time augmentation.
These superior results validate that our mapping-augmented approach provides richer and more reliable context than traditional temporal fusion, achieving better performance while using significantly fewer frames. This confirms the effectiveness of leveraging high-quality scene reconstruction priors for 3D detection.

\begin{table}[t]
\centering
\resizebox{\columnwidth}{!}{
\begin{tabular}{lcrrrr}
\toprule
{Method} & {Frames} & {AP L1} & {APH L1} & {AP L2} & {APH L2} \\
\midrule
CenterFormer~\cite{CenterFormer} & 16 & 82.3 & 80.9 & 77.6 & 76.3 \\
BEVFusion~\cite{bevfusion}  & 3 & 82.7 & 81.4 & 77.7 & 76.3 \\
MSF~\cite{Msf}  & 4 & 83.1 & 81.7 & 78.3 & 77.0 \\
FSD++~\cite{FSD}  & 7 & 83.5 & 82.1 & 78.4 & 77.1 \\
LoGoNet~\cite{logonet} & 5  & 83.1 & 81.8 & 78.4 & 77.1 \\
Octopus\_Noah & 2 & 83.1 & 81.7 & 78.7 & 77.3 \\
SEED-L~\cite{seed}  & 3   & 83.5 & 82.2 & 78.7 & 77.3 \\
LION~\cite{lion}    & 3     & 83.7 & 82.4 & 78.7 & 77.4 \\
VeuronNet3D  & 3  & 83.7 & 82.2 & 79.1 & 77.7 \\
HIAC     &  5     & 84.0 & 82.6 & 79.2 & 77.8 \\
InceptioLidar  & 10   & 83.8 & 82.5 & 79.2 & 77.8 \\
VADet~\cite{huang2025vadet} & 16  & 84.1 & 82.8 & 79.4 & 78.2 \\
MT3D  &    4        & 85.0 & 83.7 & 80.1 & 78.7 \\
LIVOX Detection & 7 & 84.8 & 83.5 & 80.2 & 79.0 \\
MAD~\cite{MAD}  & 100  & 86.0 & 84.3 & 81.8 & 80.2 \\
\rowcolor{blue!10}
MPA3D (Ours) & 4 & \textbf{87.2} & \textbf{85.9} & \textbf{83.0} & \textbf{81.6}  \\
\bottomrule
\end{tabular}
}
\caption{\textbf{Results on WOD test set leaderboard.}
Our method achieves state-of-the-art performance using a total of
4 frames (including 3 history frames). 
We use \textbf{bold numbers} to highlight the best results.}
\label{tab:wod_test}
\end{table}

\subsection{Ablation study}

\paragraph{Effectiveness of Mapping Priors.}

\begin{table*}[t]
\centering
\resizebox{0.85\textwidth}{!}{
\begin{tabular}{lcc cc cc cc cc}
\toprule
\multirow{2}{*}{Base Model} & \multirow{2}{*}{Surfel} & \multirow{2}{*}{3DGS} & \multicolumn{2}{c}{Vehicle L2} & \multicolumn{2}{c}{Pedestrian L2} & \multicolumn{2}{c}{Cyclist L2} & \multicolumn{2}{c}{Overall L2} \\
\cmidrule(lr){4-5} \cmidrule(lr){6-7} \cmidrule(lr){8-9} \cmidrule(lr){10-11}
& & & AP & APH & AP & APH & AP & APH & AP & APH \\ \midrule
SWFormer$^\dagger$ & $\boldsymbol{\sf{x}}$ & $\boldsymbol{\sf{x}}$ & 79.2 & 78.4 & 73.8 & \underline{69.3} & 68.1 & 67.3 & 73.7 & 71.6 \\
SWFormer$^\dagger$ & $\checkmark$ & $\boldsymbol{\sf{x}}$ & 79.6 & 78.8 & 73.5 & 69.0 & \underline{69.7} & \underline{68.8} & \underline{74.3} & \underline{72.2} \\
SWFormer$^\dagger$ & $\boldsymbol{\sf{x}}$ & $\checkmark$ & \textbf{80.1} & \underline{79.1} & \underline{74.3} & 69.0 & 68.3 & 67.4 & 74.2 & 71.9 \\
SWFormer$^\dagger$ & $\checkmark$ & $\checkmark$ & \underline{79.9} & \textbf{79.1} & \textbf{74.3} & \textbf{69.6} & \textbf{70.4} & \textbf{69.4} & \textbf{74.9} & \textbf{72.7} \\
\midrule \midrule
Ours-baseline & $\boldsymbol{\sf{x}}$ & $\boldsymbol{\sf{x}}$ & 83.7 & 83.1 & 79.8 & 76.3 & 81.9 & 81.0 & 81.8 & 80.1 \\
Ours-baseline & $\checkmark$ & $\boldsymbol{\sf{x}}$ & 85.3 & 84.7 & \underline{80.1} & \underline{76.7} & \underline{82.8} & \underline{81.9} & \underline{82.7} & \underline{81.1} \\
Ours-baseline & $\boldsymbol{\sf{x}}$ & $\checkmark$ & \underline{85.5} & \underline{85.0} & 80.0 & 76.4 & 82.4 & 81.5 & 82.6 & 81.0 \\
Ours-baseline & $\checkmark$ & $\checkmark$ & \textbf{86.5} & \textbf{85.9} & \textbf{80.5} & \textbf{77.0} & \textbf{82.9} & \textbf{82.0} & \textbf{83.3} & \textbf{81.7}\\
\bottomrule
\end{tabular}
}
\caption{\textbf{Ablation study on different Mapping Priors.} ``SWFormer$^\dagger$" is the reproduction of SWFormer with our implementation. The baseline models do not use mapping priors ($\boldsymbol{\sf{x}}$). For each model, the best and second best results are highlighted by \textbf{bold} and \underline{underlined}.
Results are reported on a subset of WOD validation split.
}
\label{tab:abs_prior}
\end{table*}
To demonstrate the effectiveness of mapping priors, we conduct several ablation studies and report their results in Table~\ref{tab:abs_prior}. We consider two baselines: 1) our implementation of SWFormer~\cite{SWFormer}, denoted as SWFormer$^\dagger$; and 2) ``Ours-baseline", which refers to
MPA3D without mapping priors while keeping camera and LiDAR inputs. For efficiency, we evaluate these models on a validation subset and report the AP/APH at the L2 difficulty level. We conduct experiments under three different mapping prior settings: with surfel only, with 3DGS only, and with both surfel and 3DGS.
For the weaker baseline (SWFormer$^\dagger$), adding either surfel or 3DGS provides similar performance gains (\eg, the overall L2 APH improves from 71.6\% to 72.2\% and 71.9\% respectively), though we observe a slight degradation in pedestrian detection. In contrast, for the stronger baseline (Ours-baseline), we observe consistent performance improvements over the baseline model across all categories. Moreover, for both baselines, combining surfel and 3DGS consistently leads to the best performance, as our gated fusion module effectively leverages their complementary strengths.

Conceptually, a mapping prior mostly maps out the static background (since foreground objects are removed), while detection is about finding dynamic foreground objects. The reason one helps the other is intuitive: \textbf{the map prior allows the model to better separate foreground from background.} This process of elimination is particularly helpful for distant or occluded objects. Fig.~\ref{fig:teaser} provides a clear example of this: (Left) Based on sparse LiDAR data alone, the vehicle (yellow box) is ambiguous and difficult to distinguish from the surrounding background noise. (Right) By leveraging the dense map prior, the model can instantly identify all points that align with the known static background (e.g., the road surface, curbs, and vegetation). The remaining, non-matching sparse points—even if they only form a partial outline—are now clearly highlighted as a dynamic foreground object, making the vehicle's detection (green box) significantly more reliable.

\paragraph{Effectiveness of Gated Fusion.}
\begin{table}[t]
\centering
\resizebox{1.0\columnwidth}{!}{%
\begin{tabular}{lcccccccc}
\toprule
Fusion & \multicolumn{2}{c}{Vehicle L2} & \multicolumn{2}{c}{Pedestrian L2} & \multicolumn{2}{c}{Cyclist L2} & \multicolumn{2}{c}{Overall L2} \\
\cmidrule(lr){2-3} \cmidrule(lr){4-5} \cmidrule(lr){6-7} \cmidrule(lr){8-9}
 Strategy & AP & APH & AP & APH & AP & APH & AP & APH \\ \midrule
 Sum & 80.3 & 81.1 & 71.3 & 67.7 & 73.0 & 72.2 & 75.2 & 73.4 \\
 Average & 82.8 & 82.0 & 75.6 & 71.9 & 77.9 & 77.0 & 78.7 & 77.0 \\
 Concat & 83.7 & 83.1 & 78.1 & 74.5 & 79.2 & 78.4 & 80.4 & 78.7 \\
 \rowcolor{blue!10}
 Gated  & \textbf{86.5} & \textbf{85.9} & \textbf{80.5} & \textbf{77.0} & \textbf{82.9} & \textbf{82.0} & \textbf{83.3} & \textbf{81.7} \\
\bottomrule
\end{tabular}
}
\caption{\textbf{Ablation study on different fusion strategies.} We use \textbf{bold numbers} to highlight the best results.
Results are reported on a subset of WOD validation split.
}
\label{tab:abs_fusion}
\end{table}
To validate the effectiveness of the proposed gated fusion strategy, we compare it with several common baseline methods: Sum, Average, and Concatenation (Concat). ``Sum" involves element-wise addition of per-voxel features from each modality; ``Average" computes the mean value across modalities; and ``Concat" concatenate the features along the channel dimension without element-wise gating. More details on these variants are provided in the supplementary materials.

As shown in Table~\ref{tab:abs_fusion}, our gated approach consistently outperforms all other strategies across every class and metric. Notably, ``Sum" and ``Average" achieve relatively poor performance, likely because they treat all modalities equally; this causes noise or empty features from one modality to indiscriminately corrupt the valid signals from another. In contrast, our method achieves an Overall L2 AP of 83.3\%, representing a significant improvement of 2.9\% over the next best method, ``Concat" (80.4\%). This demonstrates that our adaptive gating mechanism effectively integrates diverse modalities, leading to a substantial boost in detection performance compared to simpler, non-adaptive techniques.

\section{Conclusion} \label{sec:conclusion}
In this paper, we addressed the challenge of improving 3D object detection by replacing resource-intensive HD maps with scalable mapping priors. We introduced Mapping Priors Augmented 3D detection (MPA3D), a novel framework that utilizes automatically generated scene reconstructions, specifically surfels and 3DGS to supplement sparse sensor data. Our core contributions include a scalable data pipeline for automatic prior generation and a Gated Fusion Module to adaptively integrate these priors with LiDAR and camera data, using rich static context to enhance foreground object detection. A mixed-modality training strategy facilitates robustness when priors are unavailable. Extensive experiments on WOD demonstrate that our method achieves new state-of-the-art results, surpassing both multi-frame and complex temporal-fusion methods while using significantly fewer frames. These results validate that leveraging scalable, reconstructed scene priors is a powerful and effective strategy for enhancing 3D perception.
\newpage

\clearpage
\setcounter{page}{1}
\maketitlesupplementary

\section*{Training Details}
To fully utilize the proposed scalable scene reconstruction (Surfel and 3DGS) pipelines, we propose a three-stage training strategy for the proposed \textsc{MPA3D} approach.
In the initial pre-training stage, the model is trained on 100 million internal video sequences without mapping priors (\eg, Surfels and 3DGS). To ensure scalability, we leverage an off-board auto-labeler based on~\cite{Modar} to generate 7-DoF bounding boxes and category labels. Subsequently, we conduct a mid-training phase on a curated dataset of approximately 350K sequences. In this stage, we incorporate mapping priors (\ie, Surfels and 3DGS reconstructions), and utilize high-quality manual annotations to further refine the model. Finally, we fine-tune the model on the Waymo Open Dataset (WOD)~\cite{Waymo} training set, utilizing the full complement of mapping priors to yield the final detector.



\section*{Concatenation Fusion}
In Section 4.4 of the main paper, we investigate various fusion strategies, including summation, averaging, concatenation, and gated fusion. The first two variants are straightforward, performing simple element-wise aggregation of voxel-based features across modalities. In contrast, the concatenation strategy adopts a hierarchical approach. First, the LiDAR feature $\bar{f}_{\text{lidar}}$ and the Surfel feature $\bar{f}_{\text{surfel}}$ are concatenated along the channel dimension, yielding $f_\text{concat}=[\bar{f}_{\text{lidar}}, \bar{f}_{\text{surfel}}]$. We then employ a PointMLP to compress the channel dimension by half, projecting the features back to the size of the original LiDAR dimension. This process is repeated to integrate the 3D Gaussian features $\bar{f}_{\text{Gaussian}}$, followed by a second PointMLP for dimensionality reduction and a residual skip connection to the initial LiDAR feature.

\section*{Influence of Different Modalities}

With a smaller model MPA3D-96M, we show the effectiveness of different inputs modalities (i.e., camera, surfel, 3DGS) in Table~\ref{tab:modalities_ablation}. By integrating more more modalities, the 3D detection performance improves consistently.

\begin{table}[t!]
\small
\centering
\setlength{\tabcolsep}{4pt} 
\begin{tabular}{cccc|c} 
\toprule
\multicolumn{4}{c|}{Input Modality} & \multicolumn{1}{c}{L2 APH} \\ 
LiDAR & Camera & Surfel & 3DGS & Overall \\
\midrule
$\checkmark$     &       &    &     & 74.9      \\
$\checkmark$     & $\checkmark$    &  &        & 75.7     \\
$\checkmark$     & $\checkmark$      &  $\checkmark$    &   & 76.1     \\
$\checkmark$     & $\checkmark$    &  $\checkmark$ & $\checkmark$ & 77.4 \\
\bottomrule
\end{tabular}
\caption{Ablation study on input modalities using MPA3D-96M model. Adding surfel and 3DGS inputs leads to improved 3D detection performance in WOD validation set.}
\label{tab:modalities_ablation}
\end{table}

\section*{Runtime}
The latency of our baseline model is 245ms. When adding both mapping priors, it goes to 452ms. Note that 245ms is higher than 20ms reported in SWFormer paper, because we included extra modules for camera, and used a much larger transformer backbone without the optimized fused transformer kernels.

{
    \small
    \bibliographystyle{ieeenat_fullname}
    \bibliography{main}
}


\end{document}